\documentclass[journal]{IEEEtran}

\usepackage{cite}
\usepackage{amsmath,amssymb,amsfonts}
\usepackage{graphicx}
\usepackage{textcomp}
\usepackage{xcolor}
\usepackage{booktabs}
\usepackage{multirow}
\usepackage{hyperref}
\usepackage{cleveref}
\usepackage{balance}
\usepackage{array}
\usepackage{tabularx}
\usepackage{stfloats}
\usepackage{url}

\usepackage{algorithm}
\usepackage{algorithmic}

\usepackage{tikz}
\usetikzlibrary{shapes.geometric, arrows.meta, positioning, fit, calc, backgrounds, decorations.pathreplacing}

\definecolor{lowcolor}{RGB}{76,175,80}
\definecolor{medcolor}{RGB}{255,193,7}
\definecolor{highcolor}{RGB}{255,152,0}
\definecolor{critcolor}{RGB}{244,67,54}
\definecolor{agentblue}{RGB}{33,150,243}
\definecolor{mgmtpurple}{RGB}{156,39,176}
\definecolor{validgreen}{RGB}{76,175,80}

\tikzset{
    block/.style={rectangle, draw, fill=blue!10, text width=6em, text centered, rounded corners, minimum height=3em},
    decision/.style={diamond, draw, fill=yellow!10, text width=5em, text centered, aspect=2, inner sep=1pt},
    arrow/.style={-{Stealth[length=3mm]}, thick},
    entity/.style={rectangle, draw, fill=#1!15, text width=8em, text centered, minimum height=2.5em, rounded corners=3pt},
    msg/.style={-{Stealth[length=2.5mm]}, thick, #1},
    timeline/.style={thick, dashed, gray},
}

\def\BibTeX{{\rm B\kern-.05em{\sc i\kern-.025em b}\kern-.08em
    T\kern-.1667em\lower.7ex\hbox{E}\kern-.125emX}}

\begin{document}

\title{Criticality-Based Guard Rail Validation for AI Agent Decisions in Autonomous Telecom Networks}

\author{
\IEEEauthorblockN{Ravi Kant Sharma}\\
\IEEEauthorblockA{Ericsson\\
ravi.k.kant.sharma@ericsson.com}
}

\maketitle

\begin{abstract}
The evolution toward fully autonomous telecommunications networks (Autonomous Network Levels 4--5) requires AI/ML agents to make real-time network decisions without human intervention. However, no standardized runtime mechanism exists to intercept and validate individual inference outputs before they trigger live network state changes, creating risks of erroneous autonomous decisions. This paper proposes the Guard Rail Validation (GRV) framework, a standardizable runtime architecture for intercepting and validating AI-driven decisions before execution. The framework evaluates decisions across multiple weighted dimensions---including action scope, action type, service criticality, agent autonomy level, reversibility, and temporal behavioral patterns---to determine a criticality level. Based on this level, graduated validation mechanisms are applied: execute-with-logging, bounds checking, independent agent validation, or multi-agent consensus. The framework additionally provides cross-agent conflict detection with criticality-weighted priority resolution and runtime conformance logging for regulatory compliance (e.g., EU AI Act Article~14). We present the architecture, algorithmic procedures, O-RAN deployment model, and evaluate threat coverage against known AI/ML attacks in telecommunications.
\end{abstract}


% === INTRODUCTION ===
\section{Introduction}
\label{sec:introduction}

Modern telecommunications networks increasingly employ Artificial Intelligence (AI) and Machine Learning (ML) agents to monitor network conditions---such as traffic load, interference, and performance metrics---and to generate corresponding control decisions~\cite{3gpp_ts28105, oran_aiml_security, ericsson_blog_2025}. These agents---ranging from task-specific ML models to emerging Foundation Models capable of generalizing across diverse network scenarios~\cite{telecom_fm_2024}---operate within network management and control architectures to optimize parameters, allocate resources, and adjust configurations in near real time.

The telecommunications industry is evolving toward fully autonomous networks through phases of increasing autonomy: from AI-assisted data access, through AI-driven analysis and recommendations, to fully autonomous AI-driven operations~\cite{tmf_an_framework}. The TM Forum defines Autonomous Networks Levels 0--5, with 3GPP SA5 studying autonomous management for 6G~\cite{3gpp_sa5_6g}, O-RAN embedding AI/ML in the RAN Intelligent Controller (RIC) architecture~\cite{oran_aiml_security}, and ETSI Zero Touch Service Management (ZSM) defining zero-touch closed-loop automation.

In this landscape, AI decisions become network actions through a pipeline: (1)~an AIMLInferenceFunction runs a trained ML model analyzing network data; (2)~the model produces an inference output (e.g., ``reduce transmit power on cell gNB-DU-7/NRCellDU-12 by 3dB''); (3)~this output is consumed by a network management function which translates it into an O1 configuration change, A1 policy, or E2 control message; and (4)~the network action executes on the live network, affecting real users and services.

However, there is no standardized step between step~(2) and step~(3). If a model is poisoned, compromised, or simply produces an erroneous output, the unsafe decision executes immediately with no safety check~\cite{3gpp_ts28105}. This gap creates five critical problems:

\begin{enumerate}
    \item \textbf{No pre-action validation}: The AIMLInferenceEmulationFunction provides pre-deployment sandboxing but no runtime protection once deployed.
    \item \textbf{No criticality awareness}: Not all decisions carry equal risk---reading a PM counter versus shutting down a cell serving emergency services requires fundamentally different handling. No standard defines criticality classification for AI/ML agent decisions.
    \item \textbf{No graduated response}: The only option is binary activate/deactivate of the entire inference function (\texttt{activationStatus = DEACTIVATED}), which blocks \textit{all} inference from that function rather than a single unsafe decision.
    \item \textbf{No conflict resolution}: In multi-agent deployments, multiple agents may produce contradictory decisions for the same network entity with no standardized detection or resolution mechanism~\cite{gb2628888, ep4595478}.
    \item \textbf{No regulatory compliance}: Emerging AI regulations---such as the EU AI Act~\cite{eu_ai_act} Article~14(4)---require human oversight capabilities for high-risk AI systems, a requirement that current telecom standards do not natively fulfill via automated workflows.
\end{enumerate}

O-RAN WG11 identifies ``Output Integrity Attack'' (ML09:2023) as a HIGH-impact threat~\cite{oran_aiml_security} but leaves the functional architecture and interface specifications for application-layer mitigation undefined. Existing approaches~\cite{gb2640186, ep4677817} address related but distinct problems without providing runtime decision safety for individual inference outputs.

In this paper, we propose a \textit{Guard Rail Validation (GRV)} framework that addresses all five gaps. The key contributions are:

\begin{itemize}
    \item A \textbf{multi-dimensional criticality evaluation} with temporal pattern detection for agent behavioral anomalies (Section~\ref{sec:criticality});
    \item \textbf{Criticality-based graduated validation} applying qualitatively different mechanisms per criticality level (Section~\ref{sec:validation});
    \item \textbf{Slice-type-aware consensus thresholds} varying validation requirements by affected service criticality (Section~\ref{sec:validation});
    \item \textbf{Cross-agent conflict detection} with criticality-weighted priority resolution (Section~\ref{sec:conflict});
    \item \textbf{Runtime conformance logging} satisfying regulatory transparency requirements (e.g., EU AI Act, NIST AI RMF) (Section~\ref{sec:logging});
    \item A \textbf{Near-RT RIC variant} for latency-constrained deployments with escalation to full validation (Section~\ref{sec:oran_integration});
    \item \textbf{Protocol-level message flows} defining standardizable interfaces (Section~\ref{sec:protocol_flows}).
\end{itemize}

The remainder of this paper is organized as follows: Section~\ref{sec:related_work} reviews related work. Section~\ref{sec:architecture} presents the system architecture. Section~\ref{sec:grv_procedure} details the GRV procedure with algorithmic descriptions. Section~\ref{sec:oran_integration} describes O-RAN integration. Section~\ref{sec:protocol_flows} defines protocol-level message flows. Section~\ref{sec:policy} describes policy configuration. Section~\ref{sec:evaluation} evaluates threat coverage and regulatory compliance. Section~\ref{sec:walkthrough} presents an end-to-end walkthrough. Section~\ref{sec:discussion} discusses limitations. Section~\ref{sec:conclusion} concludes.

\subsection{Motivating Example}

Consider an energy-saving rApp that autonomously deactivates underutilized cells to reduce power consumption. The rApp correctly identifies NRCellDU-12 as having zero traffic load and issues a deactivation command. However, NRCellDU-12 is the sole cell providing EMERGENCY slice coverage in its geographic area. Without runtime validation, this individually rational decision would immediately drop emergency service capability---a catastrophic outcome from a seemingly valid optimization. This scenario motivates the need for a validation layer that considers not just whether an action is operationally valid, but whether it is \textit{safe} given the broader service context. We revisit this example in detail in Section~\ref{sec:walkthrough}.

\subsection{Research Questions}

This paper addresses two core research questions:
\begin{enumerate}
    \item[\textbf{RQ1}:] \textit{How can risk-proportional runtime validation of AI/ML agent decisions be enforced in telecommunications networks while respecting heterogeneous latency constraints across deployment tiers?}
    \item[\textbf{RQ2}:] \textit{Can a single, standardizable criticality model simultaneously satisfy operational safety requirements and regulatory transparency mandates (e.g., EU AI Act, NIST AI RMF) without imposing prohibitive overhead on routine network operations?}
\end{enumerate}

\subsection{Scope and Non-Goals}

The GRV framework is a \textit{runtime safety layer} that gates inference outputs before execution. It explicitly does \textit{not} aim to:
\begin{itemize}
    \item \textbf{Verify model correctness}: GRV does not certify that an AI model is correct or unbiased. It validates individual outputs at runtime, regardless of model quality.
    \item \textbf{Replace offline validation}: Pre-deployment testing (AIMLInferenceEmulationFunction) and model certification remain necessary. GRV complements these by providing runtime protection after deployment.
    \item \textbf{Provide formal safety guarantees}: Unlike control-theoretic safety filters or formal verification, GRV is a configurable, policy-based mechanism. It reduces risk pragmatically rather than proving safety mathematically.
    \item \textbf{Perform post-hoc anomaly detection}: GRV operates \textit{before} action execution (proactive), not after degradation is observed (reactive). It complements---but does not replace---post-deployment monitoring.
\end{itemize}

% === RELATED WORK ===
\section{Related Work}
\label{sec:related_work}

\subsection{3GPP AI/ML Management Framework}

3GPP TS~28.105~\cite{3gpp_ts28105} defines the AI/ML management framework with a dedicated set of Information Object Classes (IOCs) for the ML model lifecycle. Key entities include: \textit{AIMLInferenceFunction} (runs inference, has \texttt{activationStatus}); \textit{AIMLInferenceReport} (reports model performance with \texttt{decisionConfidenceScore}); \textit{AIMLInferenceEmulationFunction} (pre-deployment sandboxing); and \textit{AIMLManagementPolicy} (defines thresholds for ML management). However, the AIMLInferenceReport is a model performance metric---not a safety mechanism. It provides no pre-action validation, no decision blocking capability, and records no information about whether the decision was safe for the network. Once deployed, model outputs are assumed safe to consume directly.

\subsection{O-RAN Security Analysis}

O-RAN WG11~\cite{oran_aiml_security} identifies ``Output Integrity Attack'' (ML09:2023) as a HIGH-impact threat where an adversary manipulates AI/ML model output to force incorrect decisions. Model outputs in O-RAN become A1 policies, E2 control messages, or O1 configurations. The analysis recommends ``Monitoring'' as a security control but provides no specification of: what to monitor, how to evaluate output safety, what action to take on unsafe detection, or how to log decisions for audit.

\subsection{Regulatory Requirements}

The EU AI Act (Regulation 2024/1689)~\cite{eu_ai_act} categorizes AI systems by risk level. Telecommunications AI managing critical infrastructure is likely high-risk under Annex III, requiring: risk management systems (Article~9), human oversight mechanisms (Article~14), transparency and logging of decisions (Article~12), and technical robustness (Article~15). Article~14(4) specifically requires that high-risk AI systems provide the ability to ``correctly interpret the system's output,'' ``decide not to use the system or to override its output,'' and ``interrupt the system through a stop button or similar procedure.'' No existing telecom standard provides mechanisms satisfying these requirements for autonomous network management decisions.

\subsection{Existing Patent Approaches}

Table~\ref{tab:related_comparison} summarizes existing approaches and their limitations.

\begin{table*}[t]
\centering
\caption{Comparison of Existing Approaches with GRV Framework}
\label{tab:related_comparison}
\footnotesize
\begin{tabular}{@{}lllcccc@{}}
\toprule
\textbf{Approach} & \textbf{Mechanism} & \textbf{Criticality} & \textbf{Conflict} & \textbf{Regulatory Audit} & \textbf{Std.} \\
\midrule
3GPP TS 28.105~\cite{3gpp_ts28105} & Model lifecycle mgmt. & \texttimes & \texttimes & \texttimes & \checkmark \\
O-RAN WG11~\cite{oran_aiml_security} & Threat identification & \texttimes & \texttimes & \texttimes & \checkmark \\
O-RAN WG2/WG3~\cite{oran_aiml_workflow} & Parameter conflict detection & \texttimes & Platform & \texttimes & \checkmark \\
GB2640186A~\cite{gb2640186} & Trust-building, single validator & Uniform & \texttimes & \texttimes & \texttimes \\
EP4677817A1~\cite{ep4677817} & Agent exploration control & Cell-level & \texttimes & \texttimes & \texttimes \\
EP4595478A1~\cite{ep4595478} & Post-deployment rate limiting & \texttimes & \texttimes & \texttimes & \texttimes \\
US12309027B2~\cite{us12309027} & Conflict coordination & \texttimes & Basic & \texttimes & \texttimes \\
\midrule
\textbf{GRV (ours)} & \textbf{Graduated validation} & \textbf{Multi-dim.} & \textbf{Crit.-weighted} & \textbf{\checkmark} & \texttimes \\
\bottomrule
\end{tabular}
\end{table*}

EP4677817A1~\cite{ep4677817} controls which agents can perform exploratory actions based on cell criticality. However, it operates at agent-level permissioning rather than individual decision validation---it does not intercept and evaluate each inference output at runtime.

EP4595478A1~\cite{ep4595478} detects performance degradation post-deployment and modifies action rates. This is a reactive approach---by the time degradation is detected, unsafe actions have already executed. Our framework provides proactive pre-action validation.

US12309027B2~\cite{us12309027} and GB2628888B~\cite{gb2628888} provide conflict resolution mechanisms but lack criticality-based graduated validation, temporal pattern detection, and regulatory compliance logging.

\subsection{Runtime Safety in Adjacent Domains}

Runtime intervention on autonomous decisions has been studied in adjacent domains: \textit{shielded RL}~\cite{alshiekh2018safe} pre-computes safe action sets from formal specifications and restricts agent choices at runtime; \textit{control barrier functions}~\cite{ames2019cbf} enforce forward invariance of safe sets via minimally invasive control corrections; \textit{runtime verification}~\cite{leucker2009rv} monitors execution traces against safety predicates, flagging violations for enforcement layers; and \textit{LLM guardrails}~\cite{rebedea2023nemo} constrain generative AI outputs via programmable rule systems.

These approaches share the runtime intervention principle with GRV but address fundamentally different operational contexts. First, they target individual agent--environment loops rather than concurrent multi-vendor agents with conflicting optimization objectives over shared infrastructure. Second, their safety criteria are system-intrinsic (physical constraints, logical specifications, or content policies) rather than derived from operational context that changes with time-of-day, subscriber density, and regulatory classification. Third, none provide the combination of graduated response (qualitatively different validation mechanisms per criticality level), cross-agent consensus, and auditable decision logging required by telecom regulatory frameworks.

% === SYSTEM ARCHITECTURE ===
\section{System Architecture}
\label{sec:architecture}

\subsection{Overview}

\begin{figure*}[t]
\centering
\resizebox{\textwidth}{!}{%
\begin{tikzpicture}[node distance=0.4cm, >=Stealth, font=\small]

% Define box styles
\tikzset{
    stagebox/.style={rectangle, minimum width=3.8cm, minimum height=2.5cm, text centered, rounded corners=2pt, text width=3.5cm},
    sublabel/.style={font=\scriptsize, text=black!70, text width=4.5cm, align=center},
    arrowstyle/.style={-{Stealth[length=4mm, width=3mm]}, line width=1.5pt, black!50},
}

% Four main boxes
\node[stagebox, fill=agentblue!80, text=white, font=\large\bfseries] (agent) {Agent\\Decision};
\node[stagebox, fill=teal!60, text=white, font=\large\bfseries, right=1.2cm of agent] (crit) {Criticality\\Evaluation};
\node[stagebox, fill=mgmtpurple!40, text=white, font=\large\bfseries, right=1.2cm of crit] (valid) {Graduated\\Validation};
\node[stagebox, fill=orange!80, text=white, font=\large\bfseries, right=1.2cm of valid] (enforce) {Enforcement};

% Sub-labels
\node[sublabel, below=0.15cm of agent] {Agents (rApps, xApps, ML Functions)};
\node[sublabel, below=0.15cm of crit] {Multi-Dimensional Scoring};
\node[sublabel, below=0.15cm of valid] {Proportional to Risk + Conflict Resolution};
\node[sublabel, below=0.15cm of enforce] {Execute | Block | Escalate};

% Arrows
\draw[arrowstyle] (agent.east) -- (crit.west);
\draw[arrowstyle] (crit.east) -- (valid.west);
\draw[arrowstyle] (valid.east) -- (enforce.west);

% GRV boundary (red box around middle three)
\draw[red!70, thick, rounded corners=4pt] ([xshift=-0.3cm, yshift=0.4cm]crit.north west) rectangle ([xshift=0.3cm, yshift=-1.5cm]enforce.south east);

\end{tikzpicture}%
}
\caption{GRV framework overview. Agent inference outputs are evaluated across multiple weighted dimensions, subjected to graduated validation proportional to the assessed criticality, and enforced with one of three outcomes.}
\label{fig:architecture}
\end{figure*}
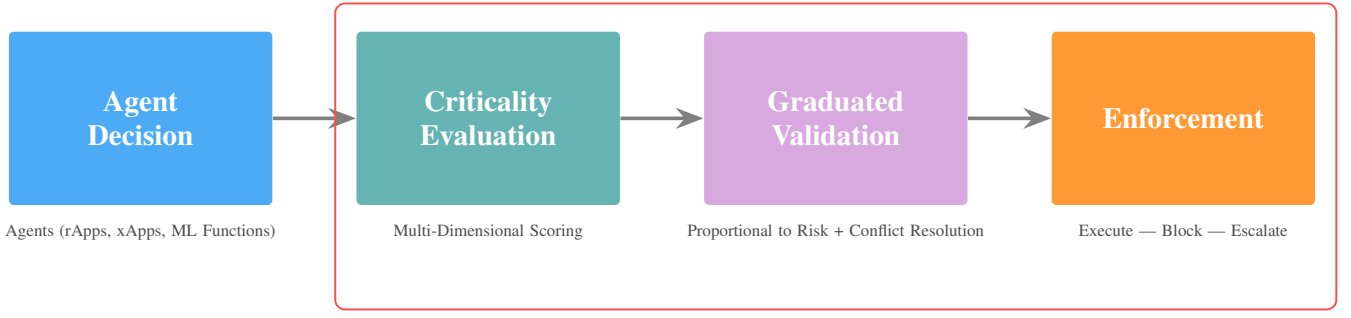

The GRV framework operates as a mandatory validation layer within the AI/ML management pipeline, positioned between inference output production and network action execution. Fig.~\ref{fig:architecture} illustrates the high-level architecture within the 3GPP/O-RAN management context. The management system---which may be part of the Service Management and Orchestration (SMO) or non-RT RIC---performs validation of inference outputs before any consumption by targeted network management functions. The GRV procedure is triggered by one of two mechanisms:

\begin{itemize}
    \item \textbf{Transparent interception}: The management system automatically captures inference outputs within the management pipeline without requiring modification of the agent.
    \item \textbf{Explicit invocation}: The AIMLInferenceFunction sends a \texttt{guardRailValidationRequest} to the management system before releasing its output.
\end{itemize}

Transparent interception requires no change to AIMLInferenceFunction behavior---only the management system's output handling path is modified. All inference outputs resulting in network state changes are subject to the GRV procedure. Read-only operations (telemetry queries, PM counter reads) are excluded via a formal rule: if \texttt{actionType = READ}, the criticality level is set to LOW regardless of other dimension scores.

\subsection{Entity Roles}

\subsubsection{Primary Agents}
AIMLInferenceFunction instances with \texttt{inferenceRole=PRIMARY} produce inference outputs specifying network actions. These are the standard inference-producing agents (rApps in the non-RT RIC, xApps in the Near-RT RIC, or standalone management functions).

\subsubsection{Validator Agents}
AIMLInferenceFunction instances with \texttt{inferenceRole=VALIDATOR} evaluate the safety of inference outputs produced by primary agents. A validator evaluates the inference output against its own independent knowledge---using a different ML model, a rule-based safety constraint engine, a network state lookup, or any combination---and returns a standardized three-value response:
\begin{itemize}
    \item \textbf{AGREE}: The action is assessed as safe to execute.
    \item \textbf{DISAGREE}: The action is assessed as unsafe (mandatory rationale field).
    \item \textbf{UNCERTAIN}: Cannot determine safety with sufficient confidence.
\end{itemize}

Validator outputs are consumed exclusively by the GRV procedure and are never forwarded to network management functions for execution. Outputs from VALIDATOR instances are exempt from guard-rail validation (preventing recursive loops). The VALIDATOR role is operator-provisioned only---it cannot be self-declared by an agent.

\textbf{Validator bootstrapping}: In single-vendor or resource-constrained deployments where diverse ML-based validators are unavailable, operators may deploy lightweight rule-based validators encoding domain constraints (e.g., ``never deactivate the last cell serving an EMERGENCY slice'') or reuse existing network management functions as validators (e.g., a coverage verification function). When no validators are available, the safety invariants apply: HIGH and CRITICAL actions without validator consensus MUST be ESCALATED with mandatory operator notification.

% === GRV PROCEDURE - CRITICALITY EVALUATION ===
\section{Guard Rail Validation Procedure}
\label{sec:grv_procedure}

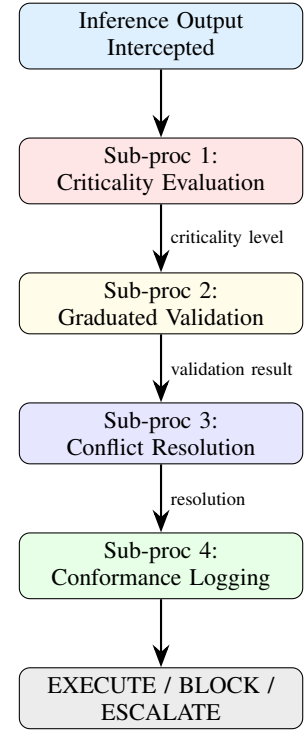
\begin{figure}[t]
\centering
\begin{tikzpicture}[node distance=0.9cm, >=Stealth, font=\small]
    \node[rectangle, draw, fill=agentblue!15, text width=10em, text centered, minimum height=2em, rounded corners] (input) {Inference Output\\Intercepted};
    \node[rectangle, draw, fill=red!10, text width=10em, text centered, minimum height=2em, rounded corners, below=of input] (sp1) {Sub-proc 1:\\Criticality Evaluation};
    \node[rectangle, draw, fill=yellow!10, text width=10em, text centered, minimum height=2em, rounded corners, below=of sp1] (sp2) {Sub-proc 2:\\Graduated Validation};
    \node[rectangle, draw, fill=blue!10, text width=10em, text centered, minimum height=2em, rounded corners, below=of sp2] (sp3) {Sub-proc 3:\\Conflict Resolution};
    \node[rectangle, draw, fill=green!10, text width=10em, text centered, minimum height=2em, rounded corners, below=of sp3] (sp4) {Sub-proc 4:\\Conformance Logging};
    \node[rectangle, draw, fill=gray!15, text width=10em, text centered, minimum height=2em, rounded corners, below=of sp4] (output) {EXECUTE / BLOCK /\\ESCALATE};

    \draw[arrow] (input) -- (sp1);
    \draw[arrow] (sp1) -- (sp2) node[midway, right, font=\scriptsize] {criticality level};
    \draw[arrow] (sp2) -- (sp3) node[midway, right, font=\scriptsize] {validation result};
    \draw[arrow] (sp3) -- (sp4) node[midway, right, font=\scriptsize] {resolution};
    \draw[arrow] (sp4) -- (output);
\end{tikzpicture}
\caption{GRV procedure overview: four sequential sub-procedures applied to each intercepted inference output.}
\label{fig:grv_overview}
\end{figure}

The GRV procedure comprises four sub-procedures executed sequentially for each intercepted inference output: (1)~Decision Criticality Evaluation, (2)~Graduated Validation, (3)~Conflict Resolution, and (4)~Runtime Conformance Logging. Fig.~\ref{fig:grv_overview} shows the overall flow.

\subsection{Sub-procedure 1: Decision Criticality Evaluation}
\label{sec:criticality}

Each inference output is evaluated across multiple weighted dimensions to determine a criticality level (LOW, MEDIUM, HIGH, or CRITICAL). The framework is domain-agnostic---dimensions are pluggable---but for autonomous telecom networks, we define the six dimensions in Table~\ref{tab:dimensions}.

\begin{table}[t]
\centering
\caption{Criticality Evaluation Dimensions}
\label{tab:dimensions}
\begin{tabular}{@{}llcc@{}}
\toprule
\textbf{Dimension} & \textbf{Categories} & \textbf{Range} & \textbf{Weight} \\
\midrule
Action Scope & Single$\to$Network-wide & 1--4 & 1.0 \\
Action Type & Read$\to$Shutdown & 0--4 & 1.0 \\
Service Criticality & mMTC$\to$Emergency & 1--4 & 1.5 \\
Autonomy Level & Advisory$\to$Autonomous & 0--2 & 1.0 \\
Reversibility & Immediate$\to$Irreversible & 0--2 & 1.0 \\
Temporal Pattern & Normal$\to$Oscillating/Burst & 0--2 & 1.5 \\
\bottomrule
\end{tabular}
\end{table}

\subsubsection{Dimension Details}

\textbf{Action Scope} assesses breadth: SINGLE\_ELEMENT~(1); MULTI\_ELEMENT~(2); SLICE~(3); NETWORK\_WIDE~(4).

\textbf{Action Type} classifies severity: READ~(0); CONFIGURE~(1); ACTIVATE~(2); DEACTIVATE~(3); SHUTDOWN~(4). The action type is determined by the management system via schema-based lookup against the NRM's attribute access types---not by the agent's self-declared type. This prevents a malicious agent from disguising state-changing actions as READ. Unknown targets default to CONFIGURE~(1).

\textbf{Service Criticality} maps from slice type: mMTC~(1); eMBB~(2); URLLC~(3); EMERGENCY~(4). When a target entity serves multiple slice types, the highest-scoring type is used. In non-sliced deployments, defaults to eMBB~(2).

\textbf{Autonomy Level} reflects operational mode: ADVISORY~(0); SUPERVISED~(1); AUTONOMOUS~(2). This is an operator-configured attribute that may change during the agent's lifecycle.

\textbf{Reversibility} assesses recoverability: IMMEDIATE~(0); DELAYED~(1); IRREVERSIBLE~(2).

\textbf{Temporal Pattern} evaluates the decision against a sliding window of recent outputs: NORMAL~(0), first decision of this type in window; FREQUENT~(1), outputs from same agent targeting same parameter exceed \texttt{frequencyThreshold} within \texttt{evaluationWindow}; OSCILLATING~(2), outputs modify same parameter in opposite directions within \texttt{oscillationWindow}; BURST~(2), outputs from $\geq$\texttt{burstThreshold} distinct agents target same entity within \texttt{evaluationWindow}. When multiple patterns are detected simultaneously, the highest score is used.

\subsubsection{Temporal State Management}

The sliding window state is maintained in-memory per (agent, parameter, target) tuple. Algorithm~\ref{alg:temporal} formalizes the temporal pattern detection logic.

\begin{algorithm}[ht]
\caption{Temporal Pattern Detection}
\label{alg:temporal}
\begin{algorithmic}[1]
\REQUIRE Output $O$, Temporal state $T$, Policy $P$
\ENSURE Temporal score $d_6 \in \{0, 1, 2\}$

\STATE $key \leftarrow (O.\text{agentDN}, O.\text{parameterName}, O.\text{targetDN})$
\STATE $window \leftarrow T.\text{getRecent}(key, P.\text{evaluationWindow})$
\STATE $score \leftarrow 0$

\COMMENT{FREQUENT: same agent, same param, same target}
\IF{$|window| + 1 > P.\text{frequencyThreshold}$}
    \STATE $score \leftarrow \max(score, 1)$
\ENDIF

\COMMENT{OSCILLATING: contradictory action within oscillation window}
\STATE $recent \leftarrow T.\text{getRecent}(key, P.\text{oscillationWindow})$
\IF{$\exists\, o \in recent$ s.t.\ $\text{contradicts}(o, O)$}
    \STATE $score \leftarrow \max(score, 2)$
\ENDIF

\COMMENT{BURST: distinct agents targeting same parameter on same entity}
\STATE $agents \leftarrow T.\text{getDistinctAgents}(O.\text{targetDN},$
\STATE $\quad O.\text{parameterName}, P.\text{evaluationWindow})$
\IF{$|agents| \geq P.\text{burstThreshold}$}
    \STATE $score \leftarrow \max(score, 2)$
\ENDIF

\STATE $T.\text{append}(key, O)$ \COMMENT{Update sliding window}
\STATE \textbf{return} $score$
\end{algorithmic}
\end{algorithm}

\textbf{Complexity}: The temporal state maintains a bounded sliding window per (agent, parameter, target) tuple. Each evaluation requires $O(W)$ operations where $W$ is the maximum number of decisions retained per window.

Legitimate coordinated operations (e.g., planned load balancing across agents) may be whitelisted in the policy to exempt them from BURST scoring.

\subsubsection{Criticality Calculation}

Algorithm~\ref{alg:criticality} presents the complete evaluation procedure.

\begin{algorithm}[ht]
\caption{Decision Criticality Evaluation}
\label{alg:criticality}
\begin{algorithmic}[1]
\REQUIRE Inference output $O$, Policy $P$, Temporal state $T$
\ENSURE Criticality level $L \in \{\text{LOW, MEDIUM, HIGH, CRITICAL}\}$

\STATE \COMMENT{Safety invariant: READ actions always LOW}
\IF{$O.\text{actionType} = \text{READ}$}
    \STATE $L \leftarrow$ LOW; \textbf{return} $L$
\ENDIF

\STATE $d_1 \leftarrow \text{evaluateScope}(O.\text{targetDN})$
\STATE $d_2 \leftarrow \text{evaluateActionType}(O, \text{NRM\_schema})$
\STATE $d_3 \leftarrow \text{evaluateSliceType}(O.\text{targetDN})$
\STATE $d_4 \leftarrow \text{getAutonomyLevel}(O.\text{agentDN})$
\STATE $d_5 \leftarrow \text{evaluateReversibility}(O.\text{actionType}, O.\text{targetDN})$
\STATE $d_6 \leftarrow \text{evaluateTemporalPattern}(O, T)$

\STATE $\mathbf{w} \leftarrow P.\text{weights}$
\STATE $S \leftarrow \sum_{i=1}^{6} w_i \cdot d_i$

\IF{$S \geq P.\text{criticalMin}$}
    \STATE $L \leftarrow$ CRITICAL
\ELSIF{$S \geq P.\text{highMin}$}
    \STATE $L \leftarrow$ HIGH
\ELSIF{$S \geq P.\text{mediumMin}$}
    \STATE $L \leftarrow$ MEDIUM
\ELSE
    \STATE $L \leftarrow$ LOW
\ENDIF
\STATE \textbf{return} $L$
\end{algorithmic}
\end{algorithm}

The weighted score $S = \sum_{i=1}^{6} w_i \cdot d_i$ has a maximum value of 21 with the default weights from Table~\ref{tab:dimensions}. Thresholds are operator-configurable via GuardRailPolicy. The MEDIUM boundary ensures that even a moderate single-element action on a critical slice triggers bounds checking; the CRITICAL boundary requires multiple high-scoring dimensions to align simultaneously.

% === GRV PROCEDURE - GRADUATED VALIDATION ===
\subsection{Sub-procedure 2: Graduated Validation}
\label{sec:validation}

\begin{figure}[ht]
\centering
\begin{tikzpicture}[>=Stealth, font=\small, node distance=0.6cm]

\tikzset{
    superbox/.style={rectangle, draw, fill=#1!15, text width=14em, minimum height=3.5em, rounded corners=3pt, font=\small, align=left, inner sep=6pt},
}

\node[superbox=lowcolor] (l1) {\textbf{LOW (Score 0--4)}\\No validation\\Outcome: EXECUTED + Log};
\node[superbox=medcolor, below=0.4cm of l1] (l2) {\textbf{MEDIUM (Score 5--8)}\\Bounds check\\Outcome: EXECUTED or BLOCKED};
\node[superbox=highcolor, below=0.4cm of l2] (l3) {\textbf{HIGH (Score 9--12)}\\Independent validator\\Outcome: EXECUTED, BLOCKED, or Fallback};
\node[superbox=critcolor, below=0.4cm of l3] (l4) {\textbf{CRITICAL (Score 13+)}\\M-of-N consensus\\Outcome: EXECUTED or ESCALATED};

\draw[-{Stealth}, thick] (l1) -- (l2);
\draw[-{Stealth}, thick] (l2) -- (l3);
\draw[-{Stealth}, thick] (l3) -- (l4);

\node[font=\scriptsize, left=0.3cm of l1, text=black!50] {Low risk};
\node[font=\scriptsize, left=0.3cm of l4, text=black!50] {High risk};

\end{tikzpicture}
\caption{Graduated validation: each criticality level triggers a qualitatively different validation mechanism.}
\label{fig:decision_tree}
\end{figure}
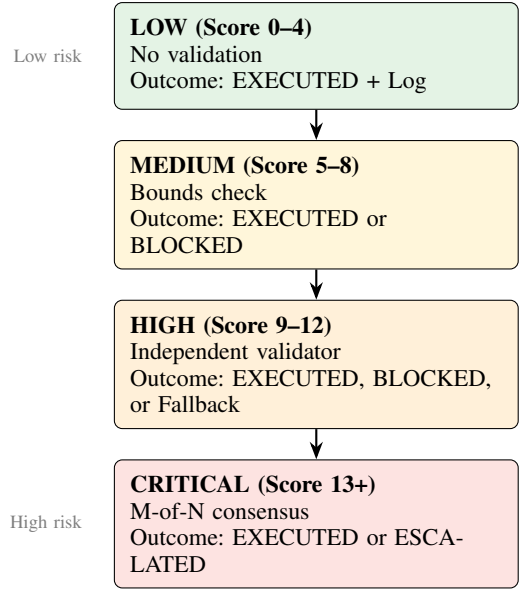

Based on the criticality level, qualitatively different validation mechanisms are applied. Fig.~\ref{fig:decision_tree} summarizes the four levels.

\subsubsection{LOW (Score 0--4): Execute with Logging}
The decision executes immediately. No additional validation is required. The conformance log records the decision for audit purposes.

\subsubsection{MEDIUM (Score 5--8): Bounds Check}
The management system checks the inference output against operator-defined safe ranges:
\begin{itemize}
    \item Is the parameter value within the allowed range? (e.g., transmit power 0--46~dBm)
    \item Is the rate of change within limits? (e.g., max 5 handover policy changes per minute)
    \item Does the action violate any hard constraints? (e.g., minimum cells per coverage area)
\end{itemize}
Bounds are defined per action type and per network entity type in the GuardRailPolicy. If no bounds definition exists for the target parameter, the check is treated as PASS with a conformance log note.

\subsubsection{HIGH (Score 9--12): Independent Validation}
The inference output is routed to a Validator AIMLInferenceFunction---an independent instance with different model architecture, different training data, or rule-based logic. On AGREE, execute. On DISAGREE, block with notification and recorded rationale. On UNCERTAIN, apply fallback policy (configurable: DELAY, REDUCE\_SCOPE, or ESCALATE). If the validator does not respond within timeout, the result is treated as UNCERTAIN.

Validators must differ in at least one of model architecture, training data provenance, or validation approach. If insufficient diverse validators are available, the procedure falls back to MEDIUM bounds check.

\subsubsection{CRITICAL (Score 13+): Multi-Agent Consensus}
The decision is routed to $N$ validators (configurable, default $N=3$). Each independently returns AGREE, DISAGREE, or UNCERTAIN. The consensus rule requires $M$-of-$N$ AGREE (default 2-of-3). For EMERGENCY slice actions, unanimous consensus is required.

The agent that produced the output is excluded from serving as its own validator. If fewer than $M$ validators respond within timeout, the procedure falls back to HIGH-level single validation. If no validators respond, the decision is BLOCKED and escalated.

% === GRV PROCEDURE - CONFLICT RESOLUTION AND LOGGING ===
\subsection{Sub-procedure 3: Conflict Resolution}
\label{sec:conflict}

After validation, the management system checks for conflicts with other pending decisions targeting the same network entity before execution.

\subsubsection{Conflict Detection}
The management system maintains a short-lived decision queue. Before executing any validated output, it checks whether another pending decision targets the same ManagedElement/NetworkSlice with contradictory intent. Two decisions conflict if they: (a)~modify the same parameter in opposite directions, or (b)~have mutually exclusive outcomes (activate vs.\ deactivate same function).

\subsubsection{Priority Evaluation and Resolution}
When conflict is detected, priority is evaluated based on the ordered criteria in Table~\ref{tab:priority}. The higher-priority decision executes; the lower-priority decision is either deferred, blocked, or escalated to the operator per policy.

\begin{table}[t]
\centering
\caption{Conflict Priority Evaluation Criteria (Ordered)}
\label{tab:priority}
\begin{tabular}{@{}clp{4.5cm}@{}}
\toprule
\textbf{Rank} & \textbf{Criterion} & \textbf{Rule} \\
\midrule
1 & Slice Criticality & EMERGENCY $>$ URLLC $>$ eMBB $>$ mMTC \\
2 & Action Urgency & Time-sensitive (fault recovery) wins \\
3 & Agent Trust & Historical reliability score \\
4 & Operator Rules & Configurable per rApp/use case \\
5 & Temporal & First-submitted wins (tie-breaker) \\
\bottomrule
\end{tabular}
\end{table}

Current conflict detection is limited to direct conflicts (same entity, same parameter). Topology-aware detection for indirect conflicts is future work.

\subsection{Sub-procedure 4: Runtime Conformance Logging}
\label{sec:logging}

Every decision passing through the GRV procedure generates a structured conformance log entry containing: agent identity, proposed action, criticality evaluation (dimension scores and level), validation result (validator responses and consensus outcome), conflict resolution if applicable, and final outcome with rationale. This satisfies regulatory transparency requirements (e.g., EU AI Act Article~14) by documenting what the system decided, how it was evaluated, and why.

% === O-RAN INTEGRATION ===
\section{O-RAN Integration}
\label{sec:oran_integration}

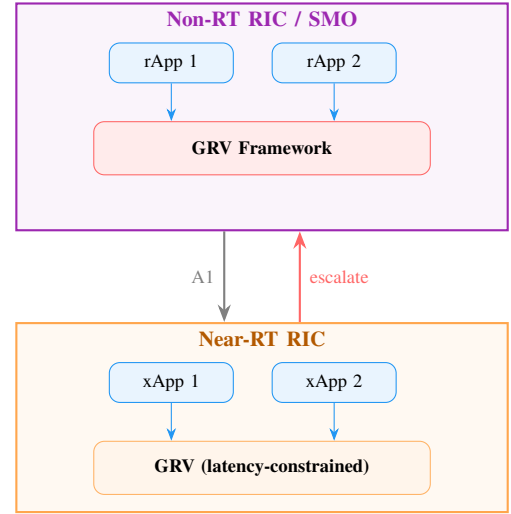
\begin{figure}[t]
\centering
\begin{tikzpicture}[node distance=0.6cm, >=Stealth, font=\small]

    % Non-RT RIC box
    \node[rectangle, draw=mgmtpurple, fill=mgmtpurple!5, thick, minimum width=6.5cm, minimum height=3cm] (nonrt) {};
    \node[anchor=north, font=\bfseries\footnotesize, text=mgmtpurple] at (nonrt.north) {Non-RT RIC / SMO};

    % rApps inside Non-RT RIC
    \node[rectangle, draw=agentblue, fill=agentblue!10, text width=4em, text centered, minimum height=1.5em, rounded corners, font=\scriptsize] at ([yshift=-0.8cm, xshift=-1.2cm]nonrt.north) (rapp1) {rApp 1};
    \node[rectangle, draw=agentblue, fill=agentblue!10, text width=4em, text centered, minimum height=1.5em, rounded corners, font=\scriptsize, right=0.5cm of rapp1] (rapp2) {rApp 2};

    % Full GRV inside Non-RT RIC
    \node[rectangle, draw=red!60, fill=red!8, text width=12em, text centered, minimum height=2em, rounded corners, font=\scriptsize, below=0.5cm of rapp1, xshift=1.2cm] (fullgrv) {\textbf{GRV Framework}};

    % Arrows: rApps -> Full GRV
    \draw[-{Stealth}, agentblue] (rapp1.south) -- (fullgrv.north -| rapp1);
    \draw[-{Stealth}, agentblue] (rapp2.south) -- (fullgrv.north -| rapp2);

    % Near-RT RIC box
    \node[rectangle, draw=orange!70, fill=orange!5, thick, minimum width=6.5cm, minimum height=2.5cm, below=1.2cm of nonrt] (nearrt) {};
    \node[anchor=north, font=\bfseries\footnotesize, text=orange!70!black] at (nearrt.north) {Near-RT RIC};

    % xApps inside Near-RT RIC
    \node[rectangle, draw=agentblue, fill=agentblue!10, text width=4em, text centered, minimum height=1.5em, rounded corners, font=\scriptsize] at ([yshift=-0.8cm, xshift=-1.2cm]nearrt.north) (xapp1) {xApp 1};
    \node[rectangle, draw=agentblue, fill=agentblue!10, text width=4em, text centered, minimum height=1.5em, rounded corners, font=\scriptsize, right=0.5cm of xapp1] (xapp2) {xApp 2};

    % Lightweight GRV inside Near-RT RIC
    \node[rectangle, draw=orange!60, fill=orange!8, text width=12em, text centered, minimum height=2em, rounded corners, font=\scriptsize, below=0.5cm of xapp1, xshift=1.2cm] (ltgrv) {\textbf{GRV (latency-constrained)}};

    % Arrows: xApps -> Lightweight GRV
    \draw[-{Stealth}, agentblue] (xapp1.south) -- (ltgrv.north -| xapp1);
    \draw[-{Stealth}, agentblue] (xapp2.south) -- (ltgrv.north -| xapp2);

    % A1 interface (policy downward)
    \draw[-{Stealth}, thick, gray] ([xshift=-0.5cm]nonrt.south) -- ([xshift=-0.5cm]nearrt.north) node[midway, left, font=\scriptsize] {A1};

    % Escalation arrow (upward from lightweight to full)
    \draw[-{Stealth}, red!60, thick] ([xshift=0.5cm]nearrt.north) -- ([xshift=0.5cm]nonrt.south) node[midway, right, font=\scriptsize, text=red!60] {escalate};

\end{tikzpicture}
\caption{O-RAN deployment model. Full GRV operates in the Non-RT RIC/SMO for rApps. A latency-constrained variant in the Near-RT RIC performs bounds checking and escalates to the full GRV on violation.}
\label{fig:oran_deployment}
\end{figure}

The GRV framework supports two deployment modes within the O-RAN architecture, differentiated by latency constraints. Fig.~\ref{fig:oran_deployment} illustrates the deployment model.

\subsection{Non-RT RIC: Full Validation}

Within the SMO, the full GRV procedure applies to rApp inference outputs before they become A1 policies or O1 configuration changes. All dimensions are evaluated, all validation levels are available, temporal pattern detection operates synchronously, and conflict detection spans all rApps managed by the SMO.

\subsection{Near-RT RIC: Latency-Constrained Variant}

The Near-RT RIC operates under strict latency constraints. This prevents full GRV evaluation (which requires external validator calls for HIGH/CRITICAL actions). The lightweight variant is a deliberate trade-off: reduced safety coverage in exchange for latency compliance.

\begin{itemize}
    \item \textbf{Bounds check only}: Actions are validated against a pre-computed local policy cache. No external validator calls.
    \item \textbf{Escalation}: Actions exceeding bounds or matching no whitelist entry are escalated to the Non-RT RIC for full GRV evaluation via the A1 interface.
    \item \textbf{Asynchronous temporal monitoring}: Pattern detection runs post-execution. This is a compensating control---it cannot prevent unsafe actions but can detect patterns and trigger operator alerts for follow-up action.
\end{itemize}

This variant does not provide equivalent safety to the full GRV. It provides best-effort pre-execution bounds checking within latency constraints, with post-execution monitoring as a safety net.

% === PROTOCOL FLOWS ===
\section{Protocol-Level Message Flows}
\label{sec:protocol_flows}

\begin{figure}[t]
\centering
\begin{tikzpicture}[font=\scriptsize, >=Stealth, node distance=0.5cm]
    % Entities
    \node[entity=agentblue, text width=4em, font=\scriptsize] (agent) {Agent};
    \node[entity=mgmtpurple, text width=3em, font=\scriptsize, right=1.5cm of agent] (mgmt) {GRV};
    \node[entity=validgreen, text width=5em, font=\scriptsize, right=1.5cm of mgmt] (v1) {Validators\\(1..N)};

    % Timelines
    \foreach \n in {agent, mgmt, v1} {
        \draw[timeline] (\n.south) -- ++(0,-7);
    }

    % 1. Inference output
    \draw[msg=agentblue] ([yshift=-1cm]agent.south) -- ([yshift=-1cm]mgmt.south)
        node[midway, above, font=\tiny] {inferenceOutput};

    % 2. Criticality evaluation
    \node[rectangle, draw=red!50, fill=red!5, font=\tiny, text width=5.5em, text centered] at ([yshift=-1.8cm]mgmt.south) (eval) {Evaluate\\$\rightarrow$ CRITICAL};

    % 3. Validation request
    \draw[msg=mgmtpurple] ([yshift=-2.8cm]mgmt.south) -- ([yshift=-2.8cm]v1.south)
        node[midway, above, font=\tiny] {validationRequest};

    % 4. Evaluate
    \node[rectangle, draw=validgreen!50, fill=validgreen!5, font=\tiny] at ([yshift=-3.6cm]v1.south) {evaluate};

    % 5. Responses
    \draw[msg=validgreen] ([yshift=-4.4cm]v1.south) -- ([yshift=-4.4cm]mgmt.south)
        node[midway, above, font=\tiny] {AGREE / DISAGREE};

    % 6. Consensus
    \node[rectangle, draw=mgmtpurple!50, fill=mgmtpurple!5, font=\tiny, text width=6em, text centered] at ([yshift=-5.3cm]mgmt.south) {Consensus:\\M-of-N AGREE?};

    % 7. Response to agent
    \draw[msg=black!70, thick] ([yshift=-6.3cm]mgmt.south) -- ([yshift=-6.3cm]agent.south)
        node[midway, above, font=\tiny] {outcome + log};

\end{tikzpicture}
\caption{CRITICAL-level consensus flow. GRV routes the inference output to $N$ validators, evaluates $M$-of-$N$ consensus, and returns the outcome.}
\label{fig:flow_critical}
\end{figure}
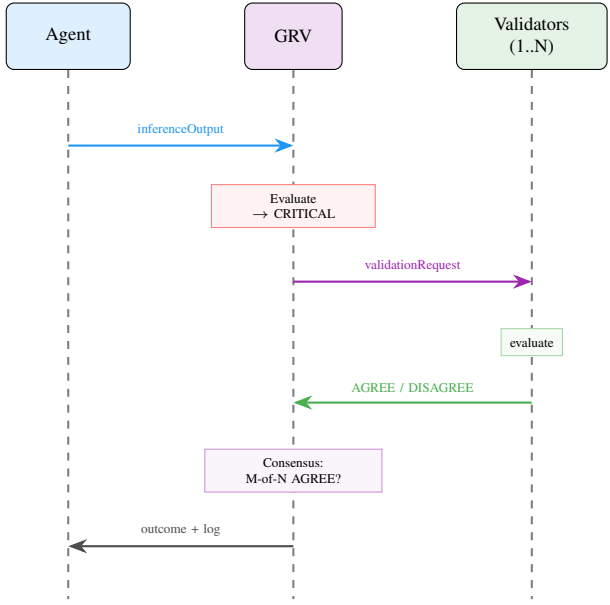

Fig.~\ref{fig:flow_critical} illustrates the CRITICAL-level flow as the most comprehensive example.

\subsection{Flow Summary}

Four protocol flows cover the GRV operating modes:

\textbf{MEDIUM (Bounds Check):} Agent produces output; GRV evaluates criticality; performs bounds check against policy; forwards if within bounds, blocks if exceeded. No external calls.

\textbf{HIGH (Independent Validation):} GRV routes output to a single diverse validator; validator returns AGREE/DISAGREE/UNCERTAIN; GRV acts accordingly.

\textbf{CRITICAL (Multi-Agent Consensus):} GRV routes output to $N$ validators (excluding the producing agent); collects responses; evaluates $M$-of-$N$ consensus; forwards or blocks with escalation.

\textbf{Explicit Invocation:} For guard-rail-aware agents that send a validation request before releasing their output. Identical evaluation logic; supports agents designed for GRV integration.

% === POLICY CONFIGURATION ===
\section{Policy Configuration}
\label{sec:policy}

All GRV parameters---criticality thresholds, dimension weights, consensus requirements, temporal windows, and conflict resolution rules---are operator-configurable via a GuardRailPolicy. Policies can be scoped globally, per slice type, per agent, or per action type, with the most specific scope taking precedence.

Policy modifications require operator authentication and are themselves logged in the conformance record, preventing unauthorized weakening of safety thresholds.

% === EVALUATION ===
\section{Evaluation}
\label{sec:evaluation}

We evaluate the GRV framework against two criteria: (1)~coverage of known AI/ML security threats in telecommunications, and (2)~alignment with EU AI Act requirements for high-risk AI systems.

\subsection{Threat Coverage Analysis}
\label{sec:threats}

Table~\ref{tab:threats} maps known AI/ML threats in telecommunications (from O-RAN WG11~\cite{oran_aiml_security}) to GRV framework defense mechanisms.

\begin{table*}[b]
\centering
\caption{Threat Coverage: GRV Defense Mechanisms Against Known AI/ML Attacks in Telecommunications}
\label{tab:threats}
\footnotesize
\begin{tabular}{@{}p{3cm}p{4cm}p{4.5cm}p{3.5cm}@{}}
\toprule
\textbf{Threat} & \textbf{Attack Vector} & \textbf{GRV Defense} & \textbf{Mechanism} \\
\midrule
Output Integrity Attack & Adversary manipulates model output to force incorrect decisions & Bounds checking catches out-of-range values; independent validation catches semantically unsafe outputs & MEDIUM: bounds; HIGH: validator \\
\midrule
Model Poisoning & Corrupted training data causes systematically bad outputs & Temporal pattern detection catches collectively destructive sequences & Temporal: FREQUENT, OSCILLATING \\
\midrule
Coordinated Multi-Agent Attack & Multiple compromised agents simultaneously target same entity & BURST detection triggers when distinct agents target same parameter within window & Temporal: BURST \\
\midrule
Action Type Spoofing & Malicious agent declares state-changing action as READ & Schema-based action type classification ignores agent's self-declared type & Criticality evaluation \\
\midrule
Oscillation Attack & Agent alternates contradictory actions to degrade network & OSCILLATING pattern detection catches parameter reversals & Temporal: OSCILLATING \\
\midrule
Privilege Escalation via Self-Validation & Agent validates its own outputs & Producing agent excluded from validator pool for its own outputs & Consensus: conflict-of-interest rule \\
\bottomrule
\end{tabular}
\end{table*}

\subsection{EU AI Act Compliance Mapping}

Table~\ref{tab:eu_ai_act} maps EU AI Act requirements for high-risk AI systems to GRV framework features.

\begin{table*}[b]
\centering
\caption{EU AI Act Requirement Mapping to GRV Framework Features}
\label{tab:eu_ai_act}
\footnotesize
\begin{tabular}{@{}p{1.5cm}p{3.5cm}p{5.5cm}p{4cm}@{}}
\toprule
\textbf{Article} & \textbf{Requirement} & \textbf{GRV Feature} & \textbf{Mechanism} \\
\midrule
Art.~9 & Risk management system & Multi-dimensional criticality evaluation with operator-configurable thresholds & Continuous risk assessment per decision \\
Art.~12 & Record-keeping and logging & Conformance log capturing full decision chain & Automated logging for every decision \\
Art.~14(4)(a) & Correctly interpret output & Criticality scores, dimension breakdown, and validator rationales & Structured decision metadata \\
Art.~14(4)(b) & Decide not to use / override output & Graduated validation: BLOCK, ESCALATE, operator notification & Autonomous override of unsafe outputs \\
Art.~14(4)(c) & Interrupt via stop button & Failsafe modes; operator notification with recommended action & Halt capability \\
Art.~15 & Accuracy, robustness, cybersecurity & Anti-bypass classification, validator diversity, temporal pattern detection & Defense-in-depth \\
\bottomrule
\end{tabular}
\end{table*}

% === WALKTHROUGH, DISCUSSION, CONCLUSION ===
\section{End-to-End Walkthrough}
\label{sec:walkthrough}

We trace an energy-saving rApp (\texttt{autonomyLevel=AUTONOMOUS}) proposing: \textit{``Deactivate NRCellDU-12 on gNB-DU-7 to save energy.''}

\begin{itemize}
    \item \textbf{Criticality Evaluation}: Schema lookup reveals NRCellDU-12 serves an EMERGENCY slice. Dimension scores: Scope=1, Type(DEACTIVATE)=3, Service(EMERGENCY)=4, Autonomy=2, Reversibility(DELAYED)=1, Temporal(NORMAL)=0. Weighted total exceeds CRITICAL threshold.

    \item \textbf{Multi-Agent Consensus}: CRITICAL level with EMERGENCY slice requires unanimous consensus. Validator-1 (rule-based): AGREE---cell load is zero. Validator-2 (ML-based): DISAGREE---sole EMERGENCY coverage in area. Validator-3 (constraint checker): DISAGREE---minimum coverage SLA violated.

    \item \textbf{Outcome}: 1-of-3 AGREE (required: 3-of-3). Decision BLOCKED. Operator notified. Conformance log written.
\end{itemize}

The EMERGENCY slice remains operational. Without GRV, the cell would have been deactivated immediately. The graduated approach ensures routine energy-saving actions proceed unimpeded while high-impact decisions receive proportionate scrutiny.

\section{Discussion}
\label{sec:discussion}

\subsection{Enabling Autonomous Network Levels 4--5}

The GRV framework provides a technical mechanism for transitioning from AN Level~3 (human-approved) to Level~4 (autonomous with guard rails)~\cite{tmf_an_framework}. Operators can delegate decisions to AI/ML agents with assurance that unsafe decisions are intercepted before execution.

\subsection{Limitations and Future Work}

\begin{itemize}
    \item \textbf{No empirical evaluation}: The framework is presented as an architecture. Performance claims (latency, distribution) require implementation and measurement against live network traffic.
    \item \textbf{False positives}: Lower thresholds increase safety but may block operationally valid actions. Threshold calibration against real operational patterns is required before deployment.
    \item \textbf{Direct conflicts only}: Current detection covers same entity, same parameter. Topology-aware detection for indirect conflicts (e.g., interference coupling) is future work.
    \item \textbf{Validator availability}: HIGH/CRITICAL levels require deployed validators. In their absence, the framework falls back to lower validation levels.
    \item \textbf{Threshold sensitivity}: Small weight changes can shift decisions between adjacent levels. Logging-only deployment for calibration is recommended before enforcement.
\end{itemize}

\section{Conclusion}
\label{sec:conclusion}

We have presented the Guard Rail Validation (GRV) framework for runtime validation of autonomous AI/ML agent decisions in telecommunications networks. The framework provides multi-dimensional criticality evaluation, graduated validation proportional to risk, temporal pattern detection, cross-agent conflict resolution, and conformance logging for regulatory compliance. The architecture is domain-agnostic in its evaluation model, with telecom-specific dimensions demonstrated as one instantiation. Future work includes prototype implementation, empirical evaluation against live network traffic, and formal standardization through 3GPP SA5.

\balance
\begingroup
\footnotesize
\bibliographystyle{IEEEtran}
\bibliography{references}
\endgroup

\end{document}